\documentclass{article} 
\usepackage[final]{colm2025_conference}

\usepackage{microtype}
\usepackage{hyperref}
\usepackage{url}
\usepackage{booktabs}
\usepackage{graphicx}

\usepackage{amsmath}
\usepackage{float}
\usepackage{algorithm}
\usepackage{algorithmic}

\usepackage{lineno}

\definecolor{darkblue}{rgb}{0, 0, 0.5}
\hypersetup{colorlinks=true, citecolor=darkblue, linkcolor=darkblue, urlcolor=darkblue}

\title{Interpreting the Latent Structure of Operator Precedence in Language Models}

\author{%
\textbf{Dharunish Yugeswardeenoo} \quad
\textbf{Harshil Nukala} \quad
\textbf{Ved Shah} \quad
\textbf{Cole Blondin} \\
\textbf{Sean O'Brien} \quad
\textbf{Vasu Sharma} \quad
\textbf{Kevin Zhu} \\
\\
Algoverse AI Research \\
3101 Park Blvd, Palo Alto, CA 94306, USA \\
\texttt{dharyugi@gmail.com, harshiln14@gmail.com, shahved25@gmail.com, cole@algoverseairesearch.org} \\
\texttt{seobrien@ucsd.edu, sharma.vasu55@gmail.com, kevin@algoverseairesearch.org}
}

\begin{document}

\ifcolmsubmission
\linenumbers
\fi

\maketitle

\begin{abstract}
Large Language Models (LLMs) have demonstrated impressive reasoning capabilities but continue to struggle with arithmetic tasks. Prior works largely focus on outputs or prompting strategies, leaving the open question of the internal structure through which models do arithmetic computation. In this work, we investigate whether LLMs encode operator precedence in their internal representations via the open-source instruction-tuned LLaMA 3.2-3B model. We constructed a dataset of arithmetic expressions with three operands and two operators, varying the order and placement of parentheses. Using this dataset, we trace whether intermediate results appear in the residual stream of the instruction-tuned LLaMA 3.2-3B model. We apply interpretability techniques such as logit lens, linear classification probes, ablations, and UMAP geometric visualization. Our results show that intermediate computations are present in the residual stream, particularly after MLP blocks. We also find that the model linearly encodes precedence in each operator's embeddings post attention layer. We introduce partial embedding swap, a technique that modifies operator precedence by exchanging high-impact embedding dimensions between operators.
\end{abstract}

\section{Introduction}

Large Language Models (LLMs) have shown impressive reasoning across a wide range of language tasks (\cite{wei2022emergentabilitieslargelanguage, chowdhery2022palmscalinglanguagemodeling, openai2024gpt4technicalreport}). Yet they are notorious for struggling with arithmetic reasoning, often producing incorrect calculations or implausible outputs (\cite{mirzadeh2024gsmsymbolicunderstandinglimitationsmathematical, bubeck2023sparksartificialgeneralintelligence}). Particularly, these errors are prominent in smaller models and remain poorly understood (\cite{gangwar2025integratingarithmeticlearningimproves,kim2024smalllanguagemodelsequation}). While recent work has shed light on how MLP layers and attention heads contribute to arithmetic reasoning, most of the studies focused on the natural language prompts and correct outputs, overlooking the processing of operator precedence and intermediate calculation steps (\cite{zhang2024interpretingimprovinglargelanguage, stolfo2023mechanisticinterpretationarithmeticreasoning, zhao2024uncoveringlargelanguagemodel}). 

We examine LLMs beyond just natural language framing and understand how arithmetic expressions are internally processed. Specifically, we focus on the model’s ability of handling order of operations in its step-by-step computation. For instance, when prompting the model to evaluate expressions like $1+1\times 2,$ does the model compute $1\times 2$ before the addition operation, or does it treat the prompt in a linear sequence irrespective of mathematical hierarchy? 

We employ a broad set of interpretability techniques including the logit lens, linear probes, ablations, partial embedding swaps, and geometric visualization via UMAP (\cite{nostalgebraist2020logitlens, alain2018understandingintermediatelayersusing, mcinnes2020umapuniformmanifoldapproximation}). All experiments are conducted on the open-source, instruction-tuned LLaMA 3.2-3B model.

\section{Related Works}
\label{gen_inst}
\subsection{Arithmetic Reasoning in Language Models}
While LLMs demonstrate strong general reasoning abilities, numerous prior works have shown persistent inconsistencies for arithmetic tasks, especially when the prompted operations require multi-step computation and manipulation (\cite{mirzadeh2024gsmsymbolicunderstandinglimitationsmathematical,bubeck2023sparksartificialgeneralintelligence, zhao2024uncoveringlargelanguagemodel}). These failures are particularly underlined in smaller models and remain poorly understood (\cite{gangwar2025integratingarithmeticlearningimproves,kim2024smalllanguagemodelsequation}). In a recent study (\cite{boye2025largelanguagemodelsmathematical}), the authors evaluated arithmetic computations across multiple different models and observed frequent inconsistencies, such as over-reliance on numerical patterns and flawed logic, even when final answers were correct. In another study (\cite{lewkowycz2022solvingquantitativereasoningproblems}), the authors have explored the use of prompting strategies, like chain-of-thought reasoning to support numerical computation. 

\subsection{Mechanistic Interpretability and Internal Components}
Research in mechanistic interpretability has focused on identifying functional components, often referred to as ``circuits", that are responsible for specific model behaviors. A circuit refers to a subnetwork of model components that offers a faithful representation of how a model solves a particular task, such as a mathematical computation (\cite{nainani2024adaptivecircuitbehaviorgeneralization}). For example, \cite{zhang2024interpretingimprovinglargelanguage} investigates the internal structure of LLMs during arithmetic tasks, showing that fine-tuning a small subset of attention heads and MLPs can enhance arithmetic performance without compromising other abilities. These components are consistent across tasks and leads to overall better arithmetic performances. (\cite{stolfo2023mechanisticinterpretationarithmeticreasoning}) adopts a causal mediation framework to trace the influence of certain internal components that contribute the most to arithmetic prediction. However, their focus is limited in scope to focus on final predictions, rather than analyzing intermediate computational steps. 

\section{Experimental Procedure}
\subsection{Model}
In this work, we utilize the open-source, instruction-tuned LLaMA 3.2-3B model (\cite{modelcard_llama3_2_3b_instruct}).
\subsection{Dataset Creation}
We constructed a synthetic dataset consisting of arithmetic expressions with three operands and two binary operators. The dataset was designed to systematically test both syntactic and semantic precedence. We consider operands $a, b, c \in \{1, 2, \dots, 9\}$ and select operator pairs $(o_1, o_2)$ from the following mixed-precedence sets:
\{ (+, *), (-, *), (+, /), (-, /) \}
Only mixed-precedence operator combinations are used to ensure meaningful precedence distinctions. For each combination of operands and operator pairs, we generate six structural variations:
\begin{enumerate}
    \item \textbf{Left-parenthesized:} $(a\ o_1\ b)\ o_2\ c$
    \item \textbf{Right-parenthesized:} $a\ o_1\ (b\ o_2\ c)$
    \item \textbf{Flipped left-parenthesized:} $(a\ o_2\ b)\ o_1\ c$
    \item \textbf{Flipped right-parenthesized:} $a\ o_2\ (b\ o_1\ c)$
    \item \textbf{No-parentheses (natural order):} $a\ o_1\ b\ o_2\ c$
    \item \textbf{No-parentheses (flipped):} $a\ o_2\ b\ o_1\ c$
\end{enumerate}
These expressions allow us to isolate the model’s handling of precedence both with and without explicit grouping via parentheses.
For simplicity, prompts were selected such that all calculations, including intermediate steps, involved only positive whole numbers. In total 8547 prompts were created, but only prompts in which the model could predict the correct answer as the top logit were used to accurately examine model behavior. The model could answer 4401 equations correctly. 
\subsection{ Logit Lens to Trace Intermediate Computation}
Before assessing how the model encodes operator precedence, we first examine whether it performs intermediate computations internally prior to generating the final output. For example, given the prompt “2 + 3 * 3 =”, we investigate whether the model computes the intermediate product 3 $\times$ 3 = 9 before arriving at the final result of 11. To account for linguistic variability in output tokens, we consider multiple surface forms (e.g., “11”, “eleven”, “eleventh”) as valid representations of the intermediate value.
Using a dataset of 4,401 prompts for which the model produces correct final answers, we utilize the logit lens technique (\cite{nostalgebraist2020logitlens}), projecting each layer's residual stream through the model’s unembedding matrix to obtain logits over the vocabulary. We then extract the top 10 tokens by logit magnitude and check whether any of them match the expected intermediate result.
\subsection{Linear Probe for Latent Intermediate Computation}
It is possible that the model performs intermediate computations internally, even when the corresponding value does not appear among the top logits. To further investigate this hypothesis, we train a linear probe (\cite{alain2018understandingintermediatelayersusing}) to predict the intermediate value directly from the model's activations at each layer, providing a complementary measure of whether such computations are linearly encoded.
\subsection{Attention Layer Ablation to Probe Causal Contribution}
To investigate whether intermediate computations are handled in the attention mechanisms of certain layers, we perform layer-wise attention ablations. For each expression, we selectively zero out the attention outputs at each layer, one layer at a time, while keeping all other components intact. For a given expression (e.g. “2 * (3 + 4) = “), we record the model’s output after ablating attention at the post-attention output before the MLP block. Ablation is implemented by setting the attention output tensor at a specific layer to zero before it is added to the residual steam. We run this procedure across the 27 transformer layers and evaluate the prediction accuracy. We then evaluate whether the intermediate result (e.g., 3 + 4 = 7) still appears in the residual stream via logit lens.

\subsection{Operator Precedence in Embedding}

To investigate whether operator precedence is encoded in specific dimensions of the model's internal representations, we introduce partial embedding swap. Consider the prompt “3 + 4 * 5 = ”, where the correct evaluation yields the answer 23, adhering to standard arithmetic precedence. However, evaluating the expression strictly left-to-right results in the incorrect answer 35. We apply the algorithm described in Appendix \ref{sec:swapAlgorithm} to probe the embedding space. We selectively swap individual dimensions between the hidden representations of the "+" and "*" operator tokens. If such a perturbation induces the model to shift its prediction from 23 to 35, it provides evidence that those dimensions contribute to encoding precedence. This intervention is performed incrementally by swapping one dimension at a time and measuring the resulting change in the logit score assigned to the token “35”. Dimensions are then ranked according to their influence on increasing this logit. In the final phase of the experiment, we perform prefix swaps of the top \textit{k} most influential dimensions to determine the minimal subset required to elevate “35” to the top-ranked prediction. This method enables us to identify specific components of the embedding space that are linked to the model’s internal representation of operator precedence.

\subsection{Low-dimensional Projection of Embeddings}
UMAP is a non-linear dimensionality reduction technique that facilitates the visualization and analysis of high-dimensional data by projecting it into a low-dimensional space (\cite{mcinnes2020umapuniformmanifoldapproximation}). 
While specific embedding dimensions appeared to correlate with operator precedence, we employed UMAP to visualize the structure of the underlying representations. To this end, we projected the activation vectors corresponding to operator tokens from a curated set of prompts. These prompts varied only in the type and position of operators, allowing for controlled comparisons. Importantly, only prompts for which the model produced the correct answer were included, ensuring that the visualizations reflected meaningful internal representations. We labeled each operator using the following format: \textit{[position of the operator] [operator name] [precedence applied to the operator]}. E.g., for a prompt  "2 *  ( 3 + 4 ) = ", the labels are "1m2" and "2p1". The multiplication sign appears first in the expression but will be evaluated second, and the plus sign appears second in the equation but will be evaluated first. 
\subsection{Logistic Probe to Identify Precedence}
This work also investigates whether operator precedence is linearly encoded in the internal representations of the model. To this end, we employ linear probes (\cite{alain2018understandingintermediatelayersusing}) trained on hidden state activations extracted before and after the first attention block (layer 0), enabling a comparative analysis of how the attention mechanism impacts the operator precendence. We construct a dataset of arithmetic prompts containing two operations whose evaluation order determines the final result. To ensure alignment between model behavior and target labels, we include only correctly answered prompts. For each valid prompt, we extract activations at operator token positions from both pre- and post-attention within layer 0. A logistic regression classifier was trained to predict whether a given arithmetic operator (e.g., +, *) corresponds to the first or second\textbf{ }operation to be evaluated in a two-step arithmetic expression (e.g., distinguishing between "( 2 + 3 ) * 4 = " and "( 2 * 3 )  + 4 = " ). Input to the probe consists of activation vectors associated with operator tokens in the prompt. The model is trained to predict either 0 or 1 based on whether the operator is evaluated first or second, respectively. For each probe, the activation vectors and binary labels are split into training and test sets (80\%/20\%). By comparing probe accuracy across the two extraction points (before and after attention), we assess whether and to what extent attention enhances the linear decodability of operator precedence.

\begin{figure}
    \centering
    \includegraphics[width=1.0\linewidth]{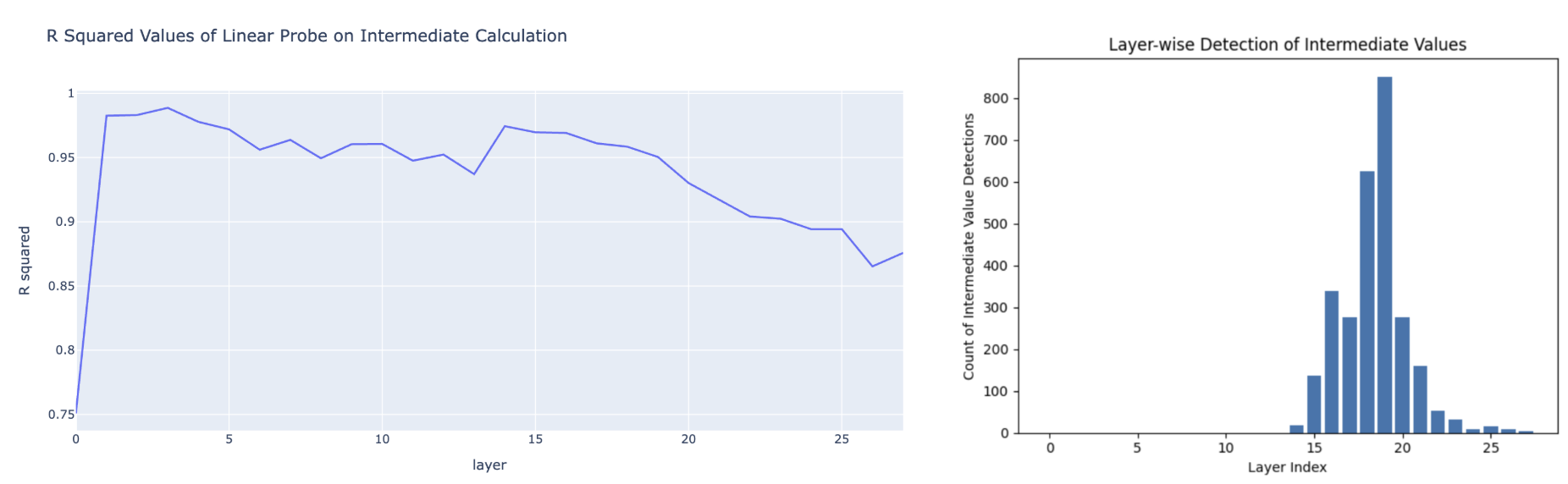}
    \caption{(Left) shows that intermediate calculations are linearly encoded in the model’s activations after layer 0, as indicated by high $R^2$ scores. (Right) Layer-wise detection frequency of intermediate computations in the residual streams at each layer. Detection peaks around layers 18–19, suggesting that the model most strongly represents intermediate computations in the later layers of its forward pass.}
    \label{fig:interm}
\end{figure}

\begin{figure}
    \centering
    \includegraphics[width=1.0\linewidth]{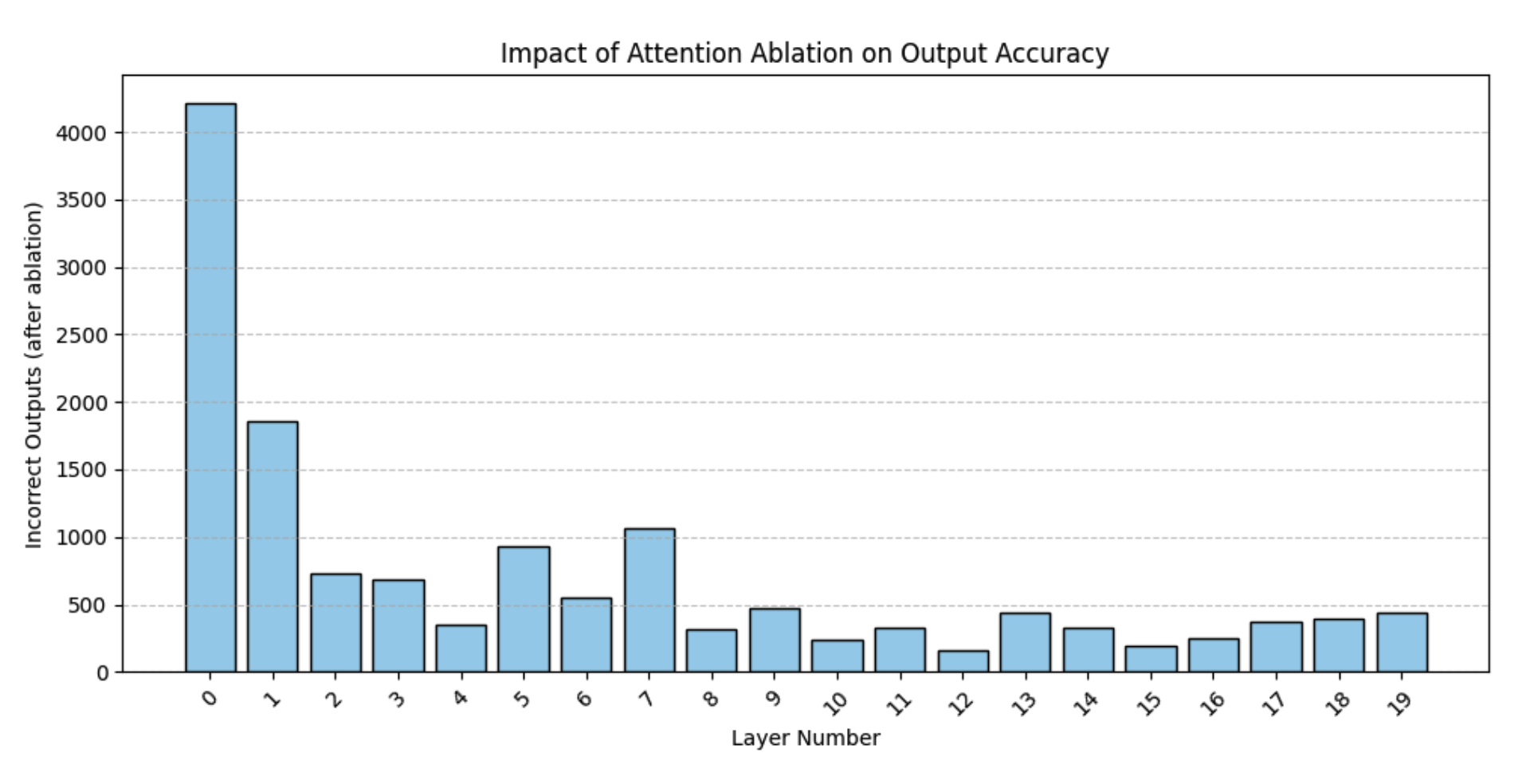}
    \caption{Shows impact of attention ablation on output accuracy across 19 layers of the model. Early layers, particularly layer 0, showed the most pronounced effect on accuracy, suggesting a key role in establishing the structural interpretation of arithmetic expressions.}
    \label{fig:ablation}
\end{figure}
\begin{figure}
    \centering
    \includegraphics[width=1.0\linewidth]{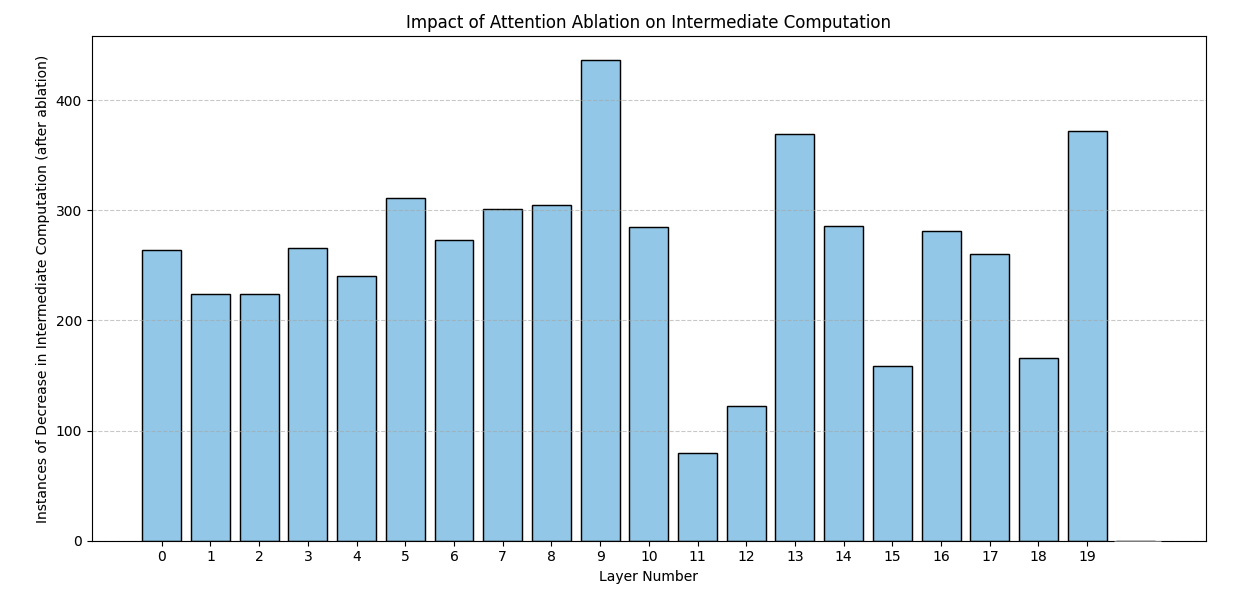}
    \caption{Shows impact of attention ablation on intermediate computation observed through Logit Lens across 4401 equations. Layers 9, 13, and 19 had the most decreases in observed intermediate values, suggesting a key role in establishing intermediate computations.}
    \label{fig:ablation+logitlens}
\end{figure}
\section{Results and Analysis}
\subsection{Intermediate Calculation}
Out of 4401 prompts, the intermediate calculation appeared 2799 times as the top logit, roughly 63.6\%. The layers in which the calculation was discovered ranged from layer 16 - 27 for the LLaMA 3.2-3B model. This distribution is shown in Figure \ref{fig:interm}. To investigate whether it is the attention block or the MLP layer that introduces the intermediate logit, we apply the unembedding matrix to the outputs of each component at the layer where the intermediate logit first appears. In all cases, we find that the intermediate answer token’s logit becomes the top-ranked logit only after the MLP block, indicating that this component is responsible for producing the intermediate computation.
Figure \ref{fig:interm} strongly suggests that the intermediate calculation is linearly encoded in the model activations after layer 0. Notably, ablating attention in early layers significantly impaired performance: ablating layer 0 led to incorrect answers in 95.7\% of cases, with layers 1, 5, and 7 also showing elevated error rates of 42.2\%, 21.1\%, and 24.2\%, respectively (Figure \ref{fig:ablation}). Layers after 19 showed limited effects. Across 19 layers, attention ablation significantly reduced the occurrence of intermediate computations observed through the logit lens, with the most pronounced effects in layers 5, 11, and 19. These findings indicate that the model relies on the attention mechanism to construct intermediate arithmetic representations throughout its depth, with layers 5, 11, and 19 playing a particularly critical role. (Figure \ref{fig:ablation+logitlens}) .
\subsection{Operator Precedence}
\begin{figure}
    \centering
    \includegraphics[width=1.0\linewidth]{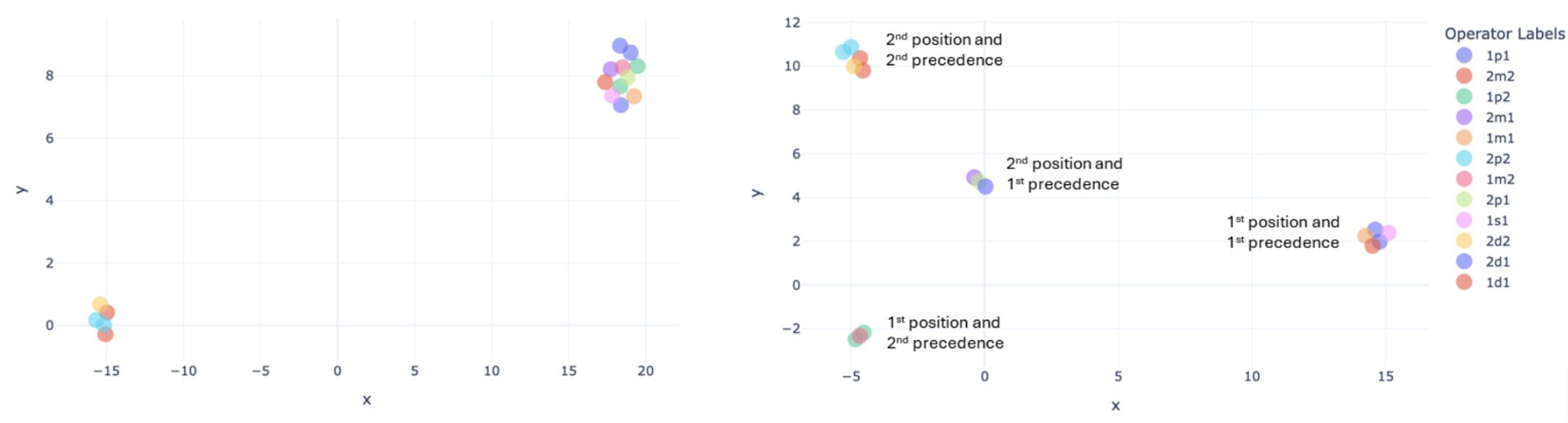}
    \caption{Low dimensional projection of operator token embeddings before (left) and after (right) the attention block in layer 0. After attention, embeddings separate by operator position and precedence, suggesting that attention encodes operator precedence information}
    \label{fig:umap-label}
\end{figure}

\begin{figure}
    \centering
    \includegraphics[width= 1.0\linewidth]{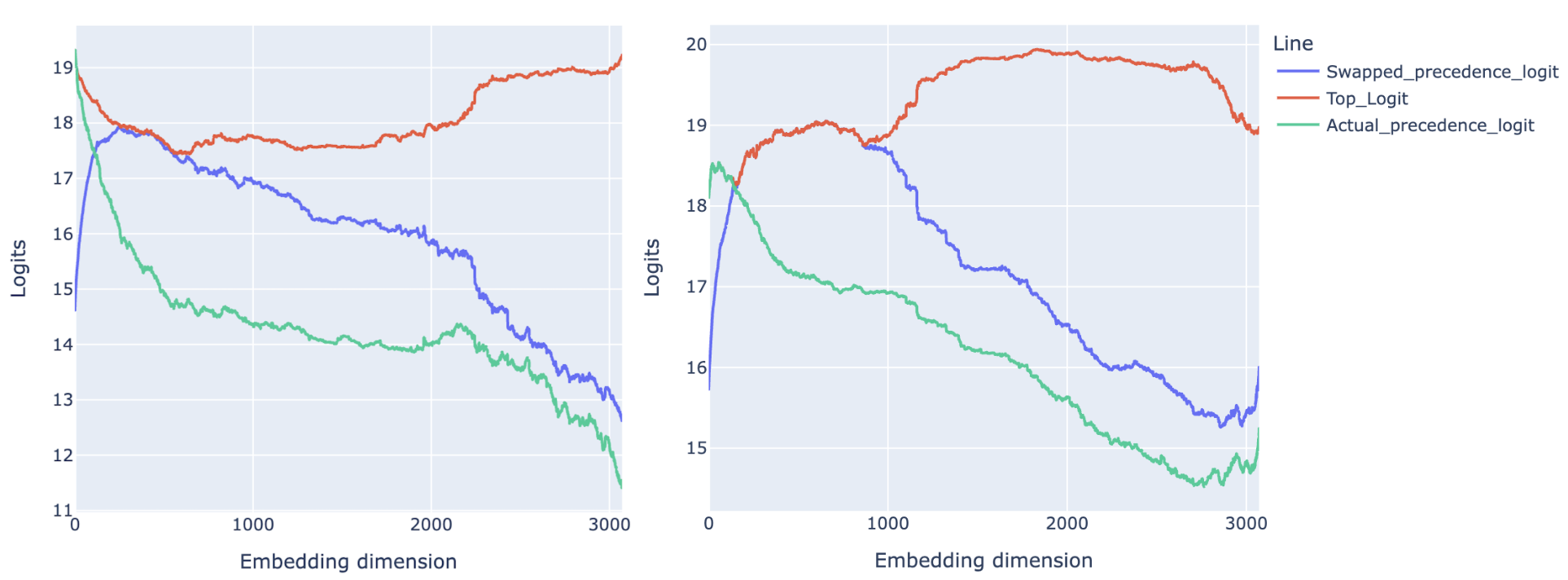}
    \caption{(left) "2 + 3 * 3 = " and (right) "4 + 8 / 4 = ". The \textit{swapped-precedence-logit} (blue) tracks the logit of the swapped (incorrect precedence) answer after patching, the \textit{actual-precedence-logit} (green) tracks the logit of the correct answer, and the \textit{top-logit} (red) shows the logit of the model's predicted token. In both examples, a subset of embedding dimensions substantially modulate the precedence-sensitive logit values, indicating that operator precedence information is sparsely localized across specific dimensions of the residual stream activations.}
    \label{fig:swap}
\end{figure}

We were able to successfully alter the model's highest logit in multiple instances using partial embedding swap, examples shown in Figure \ref{fig:swap}. 
Our UMAP projection is shown in Figure \ref{fig:umap-label}.  We found that operators who matched in both position and precedence were clustered near each other. 
Our logistic regression probe recieved 100\% accuracy on our test set, strongly indicating the presence of operator precedence.

\section*{Conclusion}
In this work, we analyze the internal representations of the LLaMA 3.2-3B model when processing arithmetic expressions involving three operands. Our findings indicate that the model performs intermediate computations internally, with such information becoming most linearly decodable in the deeper layers of the network. Through probing and intervention, we identify specific embedding dimensions that plausibly encode operator precedence, and demonstrate that modifying these dimensions via partial embedding swaps can systematically alter the model’s arithmetic predictions. Additionally, UMAP projections of operator token embeddings reveal that the model organizes operations based on both their position in the expression and their evaluation precedence.

\bibliography{colm2025_conference}
\bibliographystyle{colm2025_conference}

\appendix
\section{Appendix}

\subsection{Logit Lens Visual}
\begin{figure}[H]
    \centering
    \includegraphics[width= 1.0\linewidth]{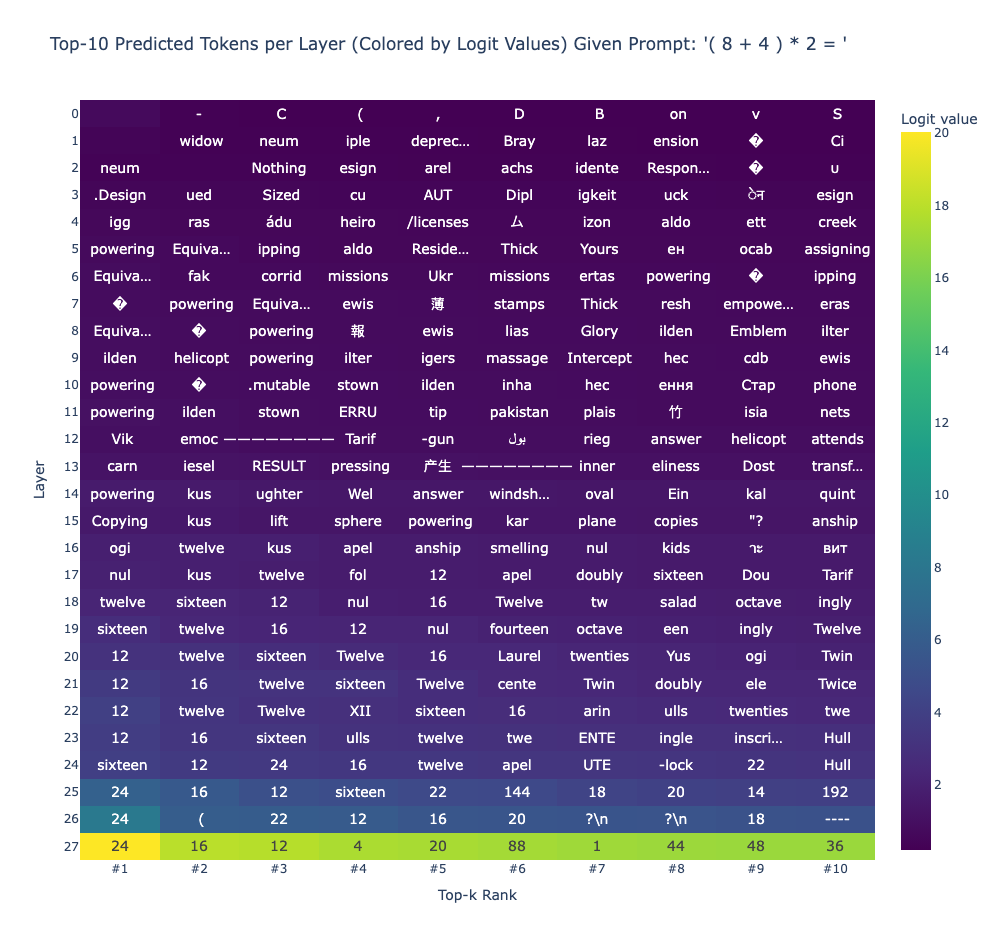}
    \caption{This figure presents the top-10 tokens ranked by their logit scores obtained via the logit lens method after each transformer layer. For the input expression under consideration, the intermediate computed value is "12". Notably, the token corresponding to "12" (or its lexical variant "twelve") emerges prior to the final output token. This observation provides evidence of the model internally representing intermediate computational steps before producing the final answer.}
    \label{fig:logitLens}
\end{figure}

\subsection{Partial Embedding Swap Algorithm}
\label{sec:swapAlgorithm}
\begin{algorithm}[t]
\caption{Identifying Influential Embedding Dimensions via Activation Swapping}
\label{alg:dim-swap}
\textbf{Input:} Prompt $P$, model $\mathcal{M}$, dimension count $d$, top-$k$ value \texttt{topk}, Operator 1 Position $Pos_1$, Operator 2 Position $Pos_2$ \\
\textbf{Output:} Sorted list of top contributing dimensions
\begin{algorithmic}[1]
\STATE Compute precedence-consistent answer token $t_{\text{target}}$ and left-to-right answer token $t_{\text{real}}$
\STATE Compute original logits $\ell^{\text{orig}} \leftarrow \mathcal{M}(P)$
\STATE Extract original logit for target token: $\ell^{\text{orig}}_{\text{target}} \leftarrow \ell^{\text{orig}}[-1, t_{\text{target}}]$
\STATE Initialize contribution vector $C \leftarrow [0] \in \mathbb{R}^d$
\FOR{each dimension index $e = 1$ to $d$}
        \STATE Define hook that swaps dimension $e$ of $Pos_1$ and $Pos_2$ tokens in the residual stream
    \STATE Compute patched logits: $\ell^{(e)} \leftarrow \mathcal{M}(P)$ with hook applied at layer 0
    \STATE Extract patched logit: $\ell^{(e)}_{\text{target}} \leftarrow \ell^{(e)}[-1, t_{\text{target}}]$
    \STATE Set contribution: $C[e] \leftarrow \ell^{(e)}_{\text{target}} - \ell^{\text{orig}}_{\text{target}}$
\ENDFOR
\STATE Rank dimensions by descending $C[e]$ and select top-\texttt{topk}
\RETURN Sorted list of top contributing dimensions and their scores
\end{algorithmic}
\end{algorithm}

\begin{algorithm}[t]
\caption{Cumulative Dimension Patching to Shift Model Prediction}
\label{alg:cumulative-patching}
\textbf{Input:} Prompt $P$, model $\mathcal{M}$, sorted contribution list $[(e_1, \Delta \ell_1), \dots, (e_d, \Delta \ell_t)]$ \\
\textbf{Output:} Number of dimensions needed to force precedence-aligned output
\begin{algorithmic}[1]
\STATE Compute original prediction $\hat{y}_{\text{orig}} \leftarrow \arg\max \mathcal{M}(P)[-1]$
\FOR{$k = 1$ to $t$}
    \STATE Define hook that swaps dimensions $e_1, \dots, e_k$ of $Pos_1$ and $Pos_2$ tokens in the residual stream
    \STATE Compute patched logits: $\ell^{(k)} \leftarrow \mathcal{M}(P)$ with hook applied at layer 0
    \STATE Extract target logit $\ell^{(k)}_{\text{target}} \leftarrow \ell^{(k)}[-1, t_{\text{target}}]$
    \STATE Compute prediction $\hat{y}^{(k)} \leftarrow \arg\max \ell^{(k)}[-1]$
    \IF{$\hat{y}^{(k)} = t_{\text{target}}$}
        \RETURN $k$ \COMMENT{Minimal number of swaps needed}
    \ENDIF
\ENDFOR
\RETURN  \COMMENT{All dimensions swapped but no change observed}
\end{algorithmic}
\end{algorithm}

\end{document}